\RequirePackage{fix-cm}
\documentclass[twocolumn]{svjour3}          
\smartqed  
\usepackage{graphicx}
\usepackage{cite}
\usepackage{amsmath,amssymb,amsfonts}
\usepackage{algorithmic}
\usepackage{textcomp}
\usepackage{subfigure} 
\newcommand{\etal}{\textit{et al.}\@ }
\newcommand{\ie}{\textit{i.e.}\@ }
\newcommand{\eg}{\textit{e.g.}\@ }
\usepackage{multirow}
\usepackage{wrapfig}
\usepackage{tabularx,booktabs}
\newcolumntype{Y}{>{\centering\arraybackslash}X}

\begin{document}

\title{Image Generation with Self Pixel-wise Normalization}


\author{Yoon-Jae Yeo \and Min-Cheol Sagong \and Seung Park \and Sung-Jea Ko \and Yong-Goo Shin }

\institute{Y.-J. Yeo \at
              School of Electrical Engineering Department, Korea University, Anam-dong, Sungbuk-gu, Seoul, 136-713, Rep. of Korea\\
              \email{yjyeo@dali.korea.ac.kr}
           \and
           M.-C. Sagong \at
              School of Electrical Engineering Department, Korea University, Anam-dong, Sungbuk-gu, Seoul, 136-713, Rep. of Korea\\
              \email{mcsagong@dali.korea.ac.kr}
           \and
           S. Park \at
              Biomedical Engineering, Chungbuk National University Hospital, 776, Seowon-gu, Cheongju-si, Chungcheongbuk-do, Rep. of Korea\\
              \email{spark.cbnuh@gmail.com}
           \and
           S.-J. Ko \at
              School of Electrical Engineering Department, Korea University, Anam-dong, Sungbuk-gu, Seoul, 136-713, Rep. of Korea\\
              \email{sjko@korea.ac.kr}
           \and
           Y.-G. Shin  \at
              Department of Artificial Intelligence, Hannam University, Daedeok-Gu, Daejeon, 34430, Rep. of Korea\\
              \email{ygshin@hnu.kr (corresponding author)}
}

\date{Received: date / Accepted: date}

\maketitle
\begin{abstract}
Region-adaptive normalization~(RAN) methods have been widely used in the generative adversarial network~(GAN)-based image-to-image translation technique. However, since these approaches need a mask image to infer the pixel-wise affine transformation parameters, they cannot be applied to the general image generation models having no paired mask images. To resolve this problem, this paper presents a novel normalization method, called self pixel-wise normalization~(SPN), which effectively boosts the generative performance by performing the pixel-adaptive affine transformation without the mask image. In our method, the transforming parameters are derived from a self-latent mask that divides the feature map into the foreground and background regions. The visualization of the self-latent masks shows that SPN effectively captures a single object to be generated as the foreground. Since the proposed method produces the self-latent mask without external data, it is easily applicable in the existing generative models. Extensive experiments on various datasets reveal that the proposed method significantly improves the performance of image generation technique in terms of Frechet inception distance~(FID) and Inception score~(IS).

\end{abstract}

\section{Introduction}
\label{sec1}
Generative adversarial network~(GAN)~\cite{goodfellow2014generative} based on convolutional neural network~(CNN) has led a series of breakthroughs for various applications including image-to-image translation~\cite{isola2017image, choi2018stargan, zhu2017unpaired, park2019semantic} and image inpainting~\cite{yu2018free, shin2020pepsi++, sagong2019pepsi}. However, due to the instability problem during the training procedure, it is still a challenge to produce high-quality images~\cite{salimans2016improved}. Since a goal of GAN is discovering the Nash equilibrium of non-convex game in the high-dimensional parameter space, GAN is substantially more complex and difficult to train compared to networks trained by supervised learning~\cite{zhang2019consistency}. To address this issue, some papers~\cite{karras2017progressive, zhang2018stackgan++, brock2018large} investigated novel network architectures for discriminator and generator. Although these methods can produce high-resolution images on challenging datasets such as ImageNet~\cite{krizhevsky2012imagenet}, they still have the fundamental problem related to the training instability.

Instead of redesigning the network architecture, many studies~\cite{zhang2019consistency, gulrajani2017improved, miyato2018spectral, roth2017stabilizing, mescheder2018training, zhou2018don, kodali2017convergence, park2021generative, yeo2021simple} attempted to penalize the discriminator for alleviating the training instability problem. The spectral normalization~(SN)~\cite{miyato2018spectral} is the most widely practiced normalization technique. In SN, the Lipschitz constraint is imposed by dividing weight matrices of the discriminator with an approximation of their largest singular value. Gulrajani~\etal~\cite{gulrajani2017improved} proposed the gradient penalty that regularizes the gradient norm of straight lines, \ie decision boundaries, between the real and generated samples. Since these normalization or regularization techniques not only mitigate the training instability problem but also effectively improve the GAN performance, most recent works apply those techniques to their applications.

On the other hand, there are only a few attempts to investigate the normalization technique for the generator. Radford~\etal~\cite{radford2015unsupervised} proposed the GAN architecture called DCGAN and empirically proved that a batch normalization~(BN)~\cite{ioffe2015batch} is effective for the generator. For a conditional GAN~(cGAN) that focuses on producing class-conditional images, Dumoulin~\etal~\cite{dumoulin2017learned} introduced a conditional batch normalization~(cBN) which performs different affine transformations according to a given condition. Brock~\etal~\cite{brock2018large} slightly modified cBN to infer the transforming parameters from not only the given class but also the latent vector. In these papers, they reveal that BN and cBN are more effective for the image generation task than other popular normalization techniques such as instance normalization~(IN)~\cite{ulyanov2016instance} and layer normalization~(LN)~\cite{ba2016layer}. Hence, most existing works adopt BN or cBN for the generator~\cite{park2021GRB, yeo2021simple, brock2018large, miyato2018cgans, miyato2018spectral, zhang2019consistency, gulrajani2017improved, roth2017stabilizing, mescheder2018training, zhou2018don, kodali2017convergence}.

Recently, some studies introduce region-adaptive normalization (RAN) techniques~\cite{yu2020region, park2019semantic, ling2021region} which perform pixel-wise affine transformations. More specifically, these methods employ the spatially-varying scaling and shifting parameters which are derived from a given mask image such as a semantic segmentation map. Although they exhibit fine performance in the field of the image-to-image translation, there is one major drawback: these approaches assume that the dataset contains reference and mask image pairs~\cite{park2019semantic}. Therefore, the existing RAN methods are not applicable to the general image generation task that does not have paired mask images. For simplicity, in the remainder of this paper, we will regard that GAN represents the image generation task. 

Indeed, most current studies~\cite{park2021GRB, park2021generative, brock2018large, miyato2018cgans, miyato2018spectral, zhang2019consistency, gulrajani2017improved, roth2017stabilizing, mescheder2018training, zhou2018don, kodali2017convergence} train the generative models using standard datasets, \eg CIFAR-10~\cite{krizhevsky2009learning}, CIFAR-100~\cite{krizhevsky2009learning}, and ImageNet~\cite{deng2009imagenet}, built for a image classification task. Since the image classification aims at classifying an object, the images of those datasets have only a single object as the foreground. Thus, a generative model trained on those datasets produces the images having the single-attributed foreground and the rest. Based on these observations, we expect that the RAN techniques can be applied to GAN if there is an extra data that discriminates between the foreground and background. However, it is not possible to pre-build the pairs of generated and mask images because we cannot predict which image will be generated.

\begin{table}[t]
\caption{Performance evaluation on CIFAR-10 dataset according to the existing normalization methods. The bold numbers indicate the best performance among the results.}
\centering
\begin{tabular}{c|c|c|c}
\hline
& BN~\cite{ioffe2015batch} & IN~\cite{ulyanov2016instance} & LN~\cite{ba2016layer} \\
\hline\hline
FID & \textbf{13.46$\pm$0.30} & 36.19$\pm$12.2 & 17.71$\pm$1.54 \\
IS & \textbf{7.77$\pm$0.02} & 6.35$\pm$0.19 & 7.48$\pm$0.12 \\
\hline
\end{tabular}
\label{table1:norm}
\end{table}

To resolve this problem, this paper presents a novel normalization method, called self pixel-wise normalization (SPN), which performs pixel-adaptive affine transformation without the external mask image. In the proposed method, SPN produces a self-latent mask that divides the input feature maps into the foreground and background regions. Then, the intrinsic transforming parameters of each region are inferred from the self-latent mask. The proposed method is simple but surprisingly effective for the image generation. To reveal the superiority of the proposed method, we conduct extensive experiments with various datasets including CIFAR-10~\cite{krizhevsky2009learning}, CIFAR-100~\cite{krizhevsky2009learning}, LSUN~\cite{yu2015lsun}, and tiny-ImageNet~\cite{yao2015tiny, deng2009imagenet}. In addition, we conduct plenty of ablation studies to prove the generalization ability of the proposed method. Quantitative evaluations show that the proposed method significantly improves the performance of both GAN and cGAN in terms of Frechet inception distance~(FID)~\cite{heusel2017gans} and Inception score~(IS)~\cite{salimans2016improved}. 

Key contributions of our paper are summarized as follows: 
\begin{itemize}
\item For the image generation task, we introduce a new approach to carry out pixel-adaptive affine transformation without the external data. Specifically, we propose a novel normalization method, called SPN, which effectively boosts the generative performance.
\item We demonstrate that SPN effectively improves the performance of both GAN and cGAN. For instance, the proposed method significantly improves FID and IS on tiny-ImageNet dataset from 35.13 and 20.12 to 28.31 and 23.35, respectively.
\item The proposed method can be easily implemented to the existing state-of-the-art generators without modifying the network architectures.
\end{itemize}

\begin{figure*}[t]
\centering
\includegraphics[width=0.95\textwidth]{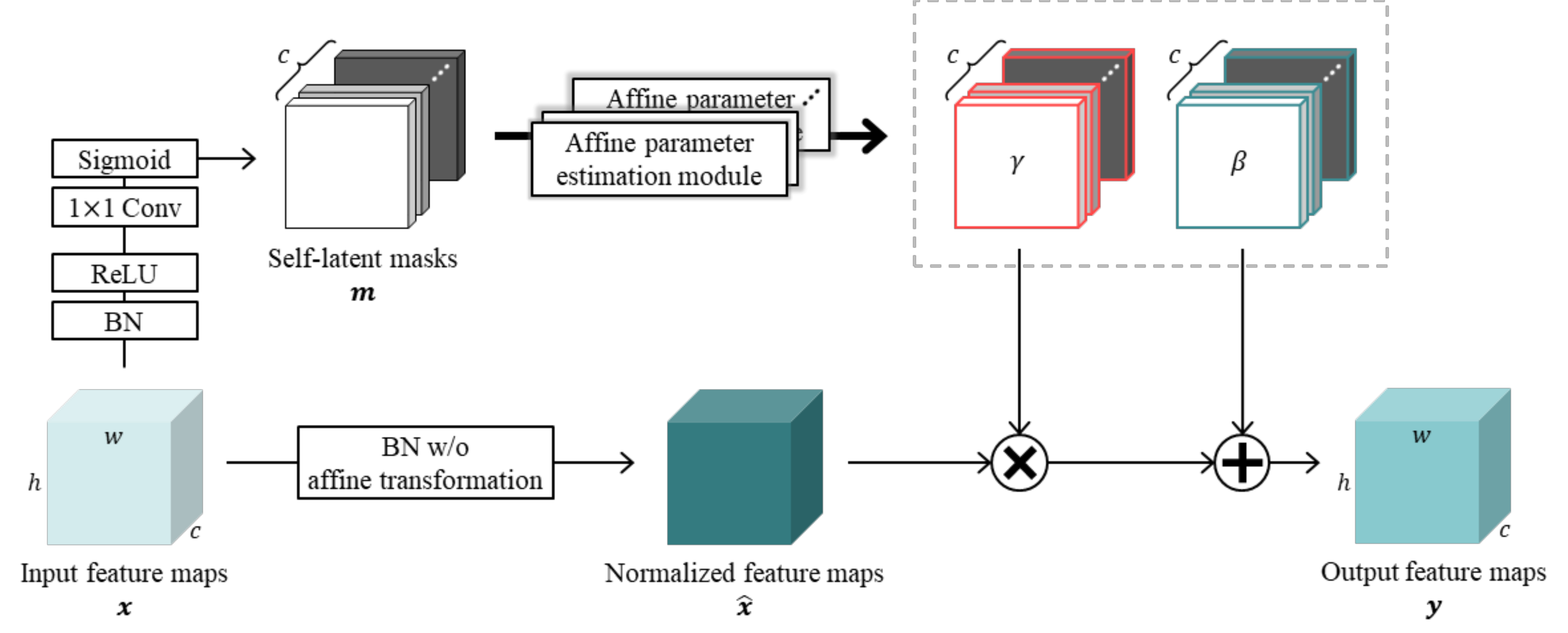}
\caption{Details of SPN. Given the intermediate feature maps, the self-latent masks are produced through few layers for each channel. Based on each mask, the scaling and shifting parameters, \ie $\gamma$ and $\beta$, are estimated for each pixel. Since these parameters are distinct in all spatial and dimensional positions, the proposed method performs the pixel-adaptive affine transformation.}
\label{fig1}
\end{figure*}

\begin{figure}[t]
\centering
\includegraphics[width=1.0\linewidth]{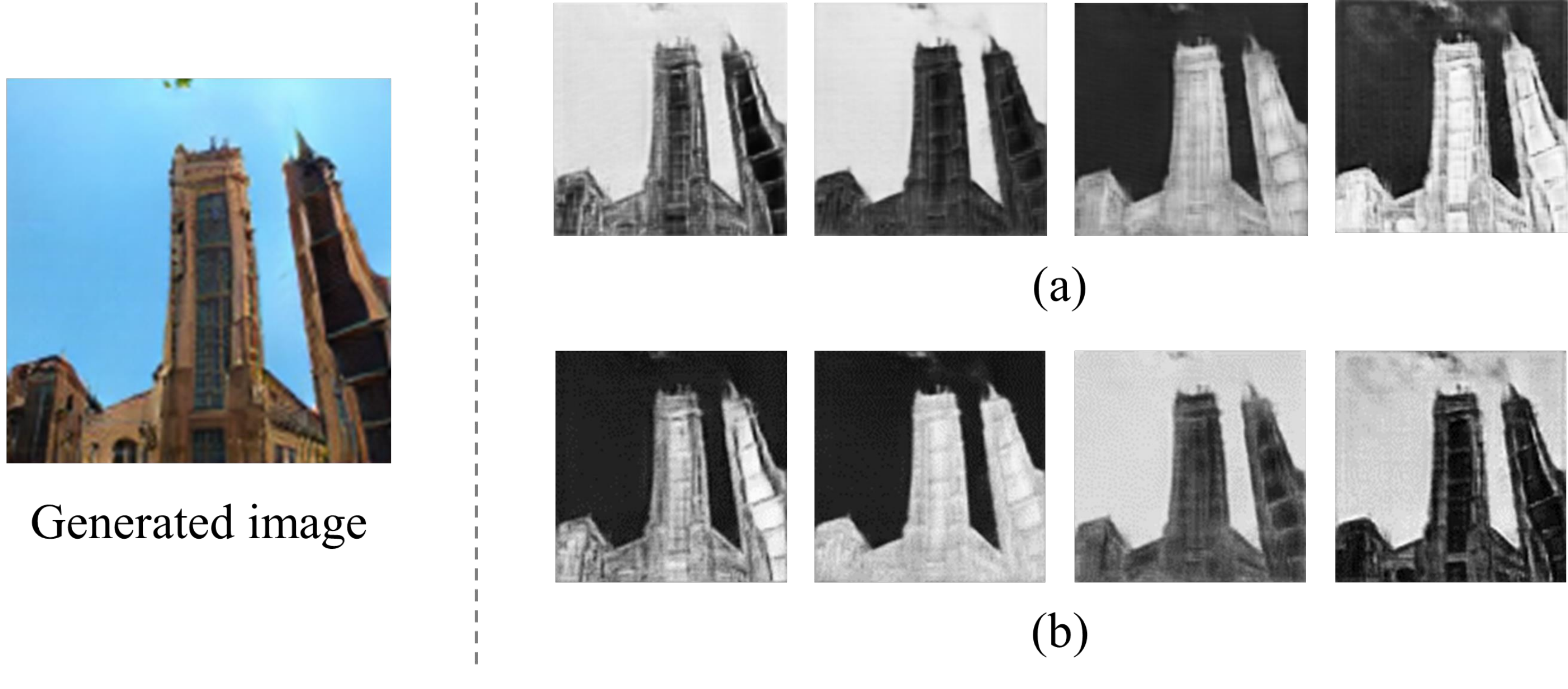}
\caption{Visualization of the self-latent masks. (a) Self latent masks $m$. (b) Inverted masks $m^*$ corresponding to the above $m$.}
\label{fig2}
\end{figure}

\section{Preliminaries}
\label{sec2}
\textbf{Generative Adversarial Network.} In the original setting, GAN~\cite{goodfellow2014generative} is composed of the generator and discriminator. In general, both networks are trained simultaneously but their objectives are contrary: the generator is trained to generate visually plausible samples, whereas the discriminator is optimized to classify real and generated ones. In order to improve the training stability and performance, many studies have been done on reformulation of the objective function. For instance, Mao~\etal~\cite{mao2017least} built the objective function using the least square errors, called least-square GAN~(LSGAN), whereas Arjovsky~\etal~\cite{arjovsky2017wasserstein} introduced Wasserstein GAN~(WGAN) that computes the loss value by measuring the Wasserstein distance between the real and generated samples. Another widely used objective function is a hinge-adversarial loss~\cite{lim2017geometric}.

On the other hand, cGAN, which aims at producing the class-conditional samples, has also been actively studied~\cite{mirza2014conditional, odena2017conditional, miyato2018cgans}. In general, cGAN employs additional conditional information, \eg class labels or text conditions, in order to control the data generation process. By optimizing the objective functions for cGAN~\cite{mirza2014conditional, miyato2018cgans}, the generator can select the image class to be generated. The reader is encouraged to review the conventional GAN and cGAN techniques for more details. 
\newline \newline 
\textbf{Normalization for GAN.} As mentioned in Section~\ref{sec1}, most existing works employ BN to their generators instead of using other normalization techniques such as IN~\cite{ulyanov2016instance} and LN~\cite{ba2016layer}. To prove that BN is more suitable for the GAN training, we compare the performance of generative models trained with BN, IN, and LN. As described in Table~\ref{table1:norm}, the generator with BN outperforms those adopting other normalization methods, and Kurach~\etal~\cite{kurach2019large} also support this. Based on these observations, we design SPN for incorporating the advantages of BN.\newline\newline
\textbf{Region-Adaptive Normalization.} The RAN methods \cite{yu2020region, park2019semantic, ling2021region} are widely used for the image-to-image translation tasks, \eg the image inpainting and the semantic image synthesis. Unlike the earlier normalization techniques such as BN and cBN, RAN requires the external data containing information differentiated by each pixel. They generally modulate the normalized feature maps by using a region-specific affine transformation whose parameters are derived from the external data. For example, Park~\etal~\cite{park2019semantic} estimate the modulating parameters from the semantic segmentation mask. However, these approaches are not applicable to the image generation task due to the absence of the mask images paired with the generated images. In other words, since the generator produces the images from the input random noise, it is hard to prepare the paired mask image. To overcome the limitation of the conventional RAN techniques, this paper presents a novel form of intrinsic normalization method specialized to the image generation. 

\section{Proposed Method}
\label{sec3}
Let $x \in \mathbb{R}^{B \times H \times W \times C}$ be a four-dimensional tensor where $B$, $H$, $W$, and $C$ indicate the size of mini-batch, height, width, and channels, respectively. 

\subsection{Self Pixel-wise Normalization}
\label{sec3.1}
Before presenting the proposed method, we briefly introduce BN~\cite{ioffe2015batch}. In BN, the input feature maps $x$ are normalized in a channel-wise manner and modulated with the learned scaling and shifting parameters. More specifically, the \textit{j}-th channel of output feature maps $y(j) \in \mathbb{R}^{B \times H \times W}$ is computed as follows: 

\begin{eqnarray}
\label{eqn:bn}
\begin{aligned}
y(j) & = \gamma_{BN}(j)\left(\frac{x(j)-\mathrm{E}[x(j)]}{\sqrt{\mathrm{Var}[x(j)]}}\right) +\beta_{BN}(j)\\
& = \gamma_{BN}(j)\widehat{x}(j)+\beta_{BN}(j), \\
\end{aligned}
\end{eqnarray}
where $\gamma_{BN}(j)$ and $\beta_{BN}(j)$ indicate scaling and shifting parameters of \textit{j}-th channel, respectively, and $\widehat{x}(j) \in \mathbb{R}^{B \times H \times W} $ is a normalized feature map. Since $\gamma_{BN}(j)$ and $\beta_{BN}(j)$ have the same value regardless of the pixel location, it is not available to conduct spatially-varied transformation on $\widehat{x}(j)$. 

\begin{figure}[t]
\centering
\includegraphics[width=0.9\linewidth]{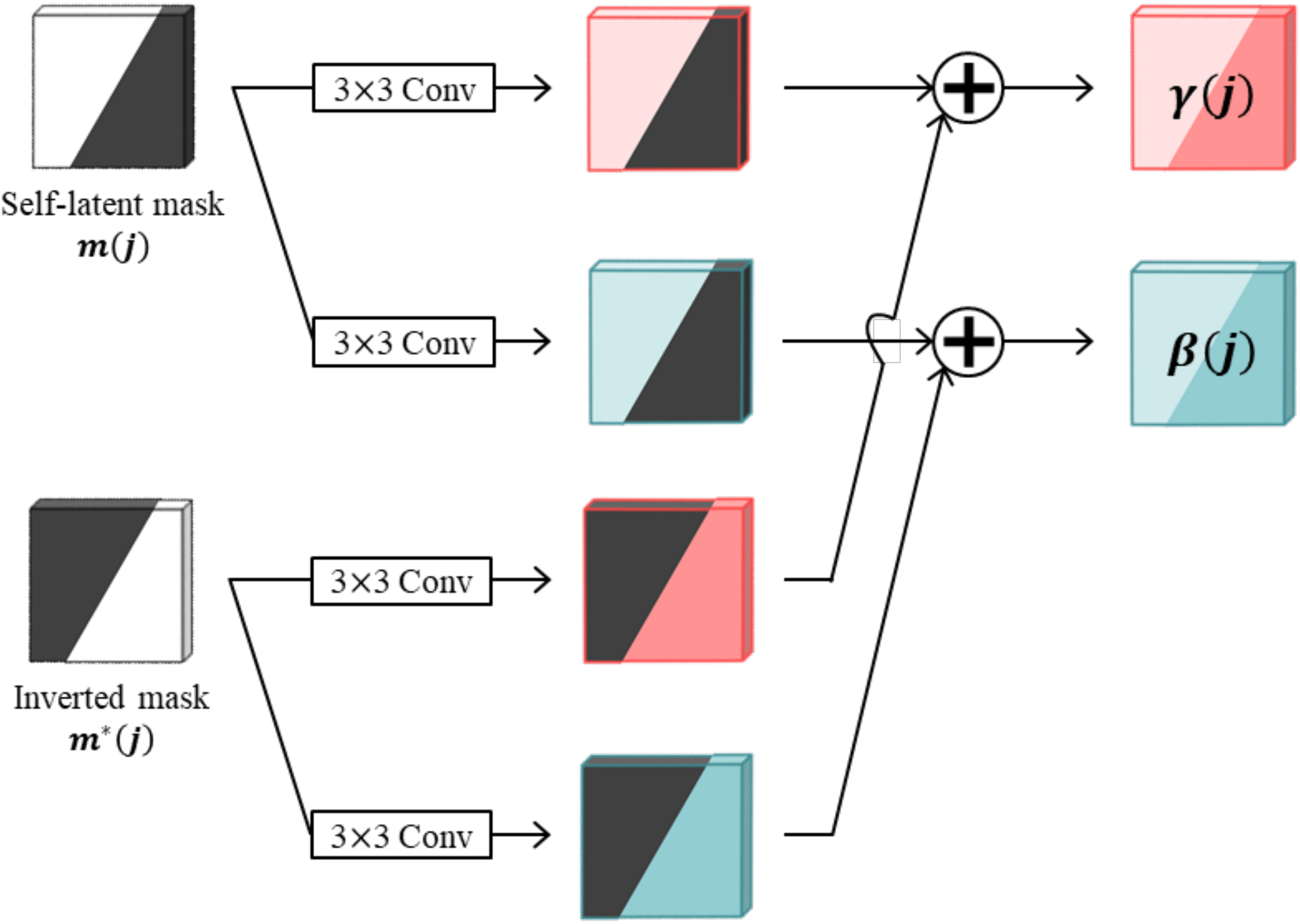}
\caption{The illustration of the affine parameter estimation module which produces $\gamma(j)$ and $\beta(j)$ using $m(j)$ and $m^*(j)$.}
\label{fig3}
\end{figure}

In contrast to BN, we design SPN to vary $\widehat{x}(j)$ with respect to the pixel location. Figure~\ref{fig1} illustrates the overall framework of SPN. In the proposed method, \textit{x} is normalized on a per-channel basis and modulated on a per-pixel basis using the learned scaling and shifting parameters. First, SPN produces self-latent masks $m \in \mathbb{R}^{B \times H \times W \times C} $ which guide the network to infer the pixel-wise transforming parameters. The key to generating $m$ in an unsupervised manner is that the generator synthesizes images composed of two types: foreground and background (see Section~\ref{sec1}). That means, if the network produces $m$ that captures two-sided regions well, the region-adaptive scaling and shifting parameters could be derived from $m$.

Based on this hypothesis, SPN draws $m$ by projecting \textit{x} onto an embedding space and passing through the sigmoid function. However, since the network is trained without a target image, it is not predictable which part of $m$ will be activated. In other words, it is ambiguous whether the active region in $m$ is the foreground. To avoid this issue, we adopt another mask $m^*$, called inverted mask, which is simply obtained by subtracting $m$ from one. Consequently, $m$ and $m^*$ are complementary to represent the two-sided regions. To clearly show the role of $m$, we present an example of visualized masks. In this example, we train the network on LSUN-church dataset and select some channels in the last SPN layer. As shown in Figure~\ref{fig2}, without the help of the external data, the network builds $m$ which precisely separates the foreground and background well. Therefore, by using $m$ and $m^*$, the scaling and shifting parameters could be estimated so as adaptive for each region.

To this end, we design an affine parameter estimation module that infers the pixel-wise transforming parameters. Since $m$ and $m^*$ have different activation maps for each channel, we derive the scaling and shifting parameters for each channel independently. That means, we employ the depth-wise convolution~\cite{howard2017mobilenets} to estimate the transforming parameters from the \textit{j}-th channel of $m$ and $m^*$. Specifically, this procedure can be formulated as follows: 
\begin{eqnarray}
\label{eqn:para}
\gamma(j) = m(j) \otimes w_{1}^{\gamma}(j) + m^*(j) \otimes w_{2}^{\gamma}(j), \\
\beta(j) = m(j) \otimes w_{1}^{\beta}(j) + m^*(j) \otimes w_{2}^{\beta}(j),
\end{eqnarray}
where $w_{i}^{\gamma}(j)$ and $w_{i}^{\beta}(j)$ represent different convolutional weights for the \textit{j}-th channel, and $\otimes$ is the convolution operation. In addition, $\gamma(j)$ and $\beta(j)$ indicate the pixel-wise scaling and shifting parameters for the \textit{j}-th channel, respectively. Note that since $\gamma(j)$ and $\beta(j)$ are inferred without affecting to other channels, the network can learn the appropriate parameters by considering the local characteristics in each channel. Figure~\ref{fig3} shows how to derive the scaling and shifting parameters from the each channel of $m$ and $m^*$. As described in Figure~\ref{fig3}, the proposed method could predict the adaptive $\gamma(j)$ and $\beta(j)$ values for each pixel. Consequently, SPN is formulated as 
\begin{eqnarray}
\label{eqn:SPN}
y(j) = \gamma(j)\widehat{x}(j)+\beta(j).
\end{eqnarray}
This equation exhibits that $\widehat{x}(j)$ is scaled and shifted by the pixel-wise affine transformation, which allows the generator to learn the region-specific features. 

\begin{figure}[t]
\centering
\includegraphics[width=1.0\linewidth]{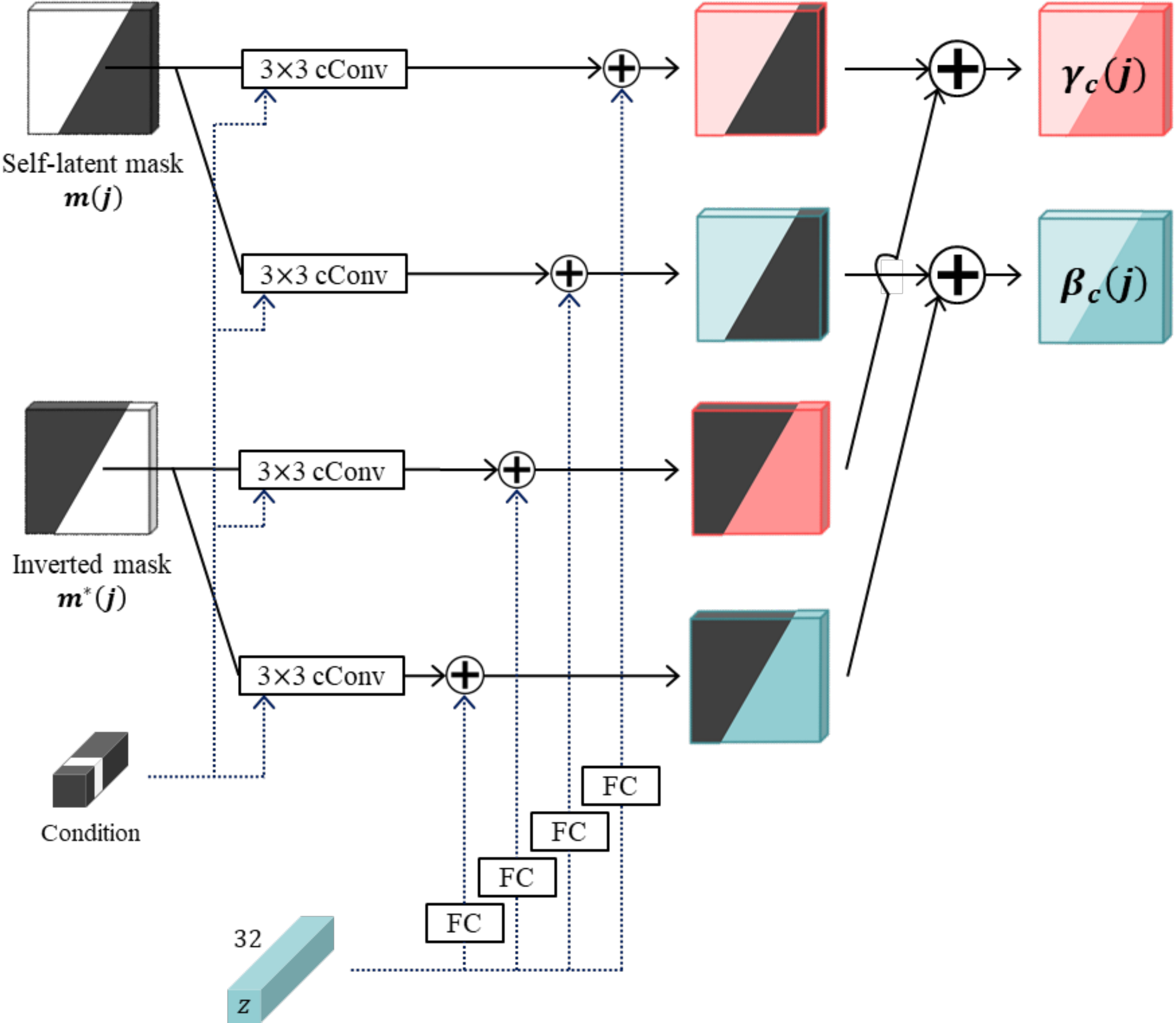}
\vspace{1mm}
\caption{The illustration of the conditional affine parameter estimation module which produces $\gamma_c(j)$ and $\beta_c(j)$ using $m(j)$ and $m^*(j)$.}
\label{fig4}
\end{figure}

\subsection{Conditional SPN}
\label{sec3.2}
In the field of cGAN~\cite{mirza2014conditional, brock2018large}, cBN~\cite{dumoulin2017learned}, which produces the different affine transformation parameters according to the given condition, is widely used. Recently, Brock~\etal~\cite{brock2018large} added the output of the fully-connected layer, which receives the noise vector as input, to allow the latent space to directly affect the affine transformation. Specifically, this approach is mathematically equivalent to adding a bias term derived from the noise vector to the scaling and shifting parameters in cBN. Since it shows better performance than the conventional cBN, we design the conditional SPN based on the method in~\cite{brock2018large}.  

In order to estimate the conditional $\gamma(j)$ and $\beta(j)$, \ie $\gamma_c(j)$ and $\beta_c(j)$, we modify the two components of SPN: the procedure of building the self-latent mask and the convolution operation in the affine parameter estimation module. First, when producing the self-latent mask, we replace BN with cBN to embed the condition-specific information. That means, we attempt to separately predict the foreground and background regions for each condition. Second, we employ a conditional convolution~(cConv)~\cite{sagong2019cgans} in the affine parameter estimation module. To provide the conditional information, cConv generates the condition-specialized weights from an embedding of class vector. Thus, owing to the conditional information in cConv, we could produce the $\gamma_c(j)$ and $\beta_c(j)$ which have different values according to the given condition. In addition, like the existing method~\cite{brock2018large}, we add the bias term derived from the noise vector. Figure~\ref{fig4} shows how the proposed method for cGAN estimates the conditional scaling and shifting parameters from each channel of $m$ and $m^*$.

\begin{table}[t]
\caption{Network architectures of the generator and discriminator for $32\times32$ (top) and $128\times128$ (bottom) images. ${*}$ indicates where the proposed method is applied.}
\centering
\vspace{2mm}

\begin{tabular}{c|c}
\hline\hline
Generator & Discriminator \\
\hline
$z \in \mathbb{R}^{128} \sim N(0, I)$ & RGB image \\
FC, $4 \times 4 \times 256$ & ResBlock, down, 128 \\
ResBlock$^{*}$, up, 256 & ResBlock, down, 128\\
ResBlock$^{*}$, up, 256 & ResBlock, 128 \\
ResBlock$^{*}$, up, 256 & ResBlock, 128 \\
BN, ReLU & ReLU \\
$3\times3$ Conv, Tanh & Global sum pooling \\
& Dense, 1 \\
\hline\hline
\end{tabular}

\vspace{0.5cm}
\begin{tabular}{c|c}
\hline\hline
Generator & Discriminator \\
\hline
$z \in \mathbb{R}^{128} \sim N(0, I)$ & RGB image \\
FC, $4 \times 4 \times 512$ & ResBlock, down, 64 \\
ResBlock, up, 512 & ResBlock, down, 128 \\
ResBlock, up, 512 & ResBlock, down, 256 \\
ResBlock$^{*}$, up, 256 & ResBlock, down, 512 \\
ResBlock$^{*}$, up, 128 & ResBlock, down, 512 \\
ResBlock$^{*}$, up, 64 & ResBlock, 512 \\
BN, ReLU & ReLU \\
$3\times3$ Conv, Tanh & Global sum pooling \\
& Dense, 1 \\
\hline\hline
\end{tabular}
\vspace{0.2cm}
\label{table:gnd}
\end{table}

\begin{figure*}[t] 
\centering
\includegraphics[width=0.95\textwidth]{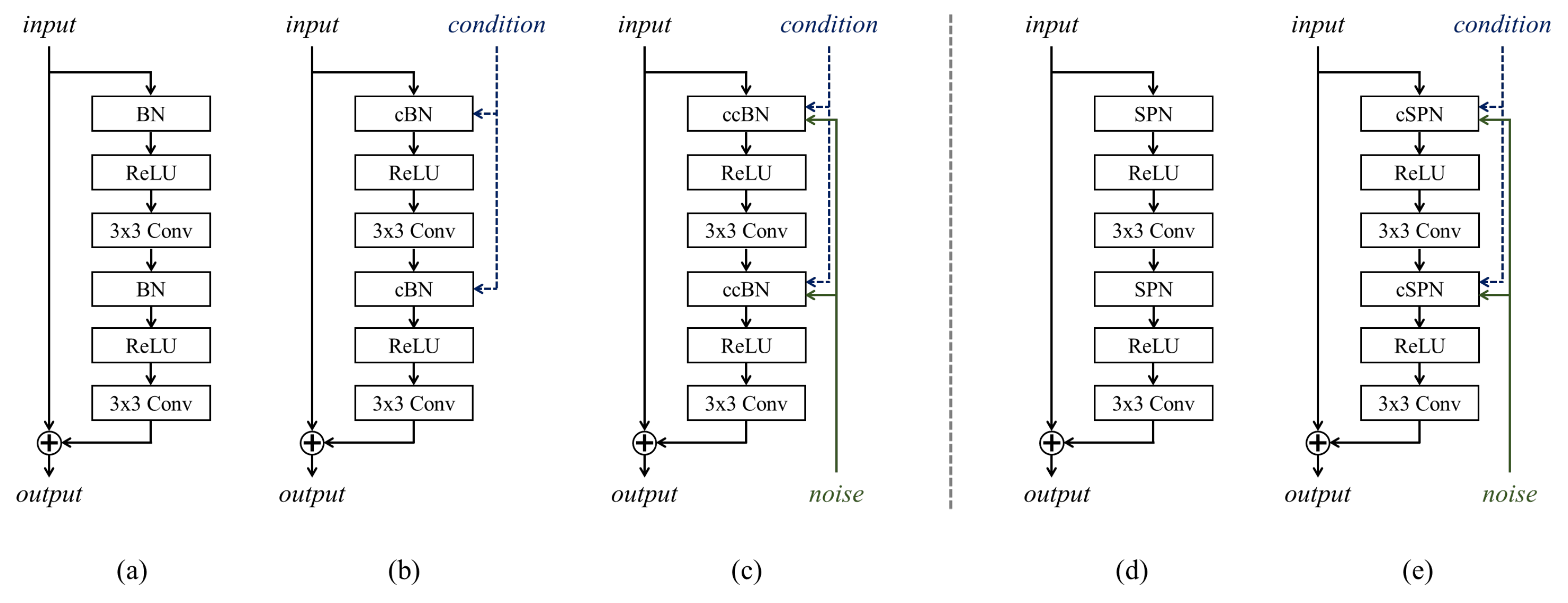}
\caption{The illustration of the residual blocks. (a) A residual block in SNGAN. (b) A residual block (\textit{condition}) in SNGAN. (c) A residual block (\textit{condition}) in BigGAN. (d) A residual block in the proposed method. (e) A residual block (\textit{condition}) in the proposed method.}
\label{sfig1}
\end{figure*}

\begin{table*}[t]
\caption{Quantitative evaluation on CIFAR-10 and CIFAR-100 datasets for GAN and cGAN. For a fair comparison, the structures are identical except for the generator's normalization technique.}
\vspace{2mm}

\centering
\begin{tabular}{c|c||cc|cc||cc|cc|cc}
\hline
&& \multicolumn{4}{c||}{GAN} & \multicolumn{6}{c}{cGAN} \\
\cline{3-12}
Dataset && \multicolumn{2}{c|}{SNGAN~\cite{miyato2018spectral}} & \multicolumn{2}{c||}{Proposed} & \multicolumn{2}{c|}{SNGAN~\cite{miyato2018spectral}} & \multicolumn{2}{c|}{BigGAN~\cite{brock2018large}} & \multicolumn{2}{c}{Proposed} \\
\cline{3-12}
&& FID & IS & FID & IS & FID & IS & FID & IS & FID & IS \\
\hline\hline
& trial1 & 13.81 & 7.79 & 12.29 & 7.90 & 10.26 & 8.06 & 9.36 & 8.09 & 7.83 & 8.28 \\
& trial2 & 13.29 & 7.76 & 11.98 & 7.94 & 10.37 & 8.02 & 9.62 & 7.97 & 7.83 & 8.35 \\
CIFAR-10 & trial3 & 13.29 & 7.76 & 12.20 & 7.94 & 10.01 & 7.99 & 9.38 & 8.02 & 7.51 & 8.41 \\
\cline{2-12}
\cite{krizhevsky2009learning} & avg. & 13.46 & 7.77 & \textbf{12.16} & \textbf{7.93} & 10.21 & 8.03 & 9.45 & 8.03 & \textbf{7.72} & \textbf{8.35} \\
& std. & 0.30 & 0.02 & 0.16 & 0.02 & 0.18 & 0.03 & 0.15 & 0.06 & 0.18 & 0.06 \\
\hline
& trial1 & 17.76 & 8.08 & 15.57 & 8.36 & 14.62 & 8.68 & 13.13 & 8.88 & 9.34 & 10.68 \\
& trial2 & 17.93 & 8.08 & 15.59 & 8.26 & 14.39 & 8.62 & 13.20 & 8.89 & 10.44 & 9.28 \\
CIFAR-100 & trial3 & 17.55 & 8.05 & 15.36 & 8.39 & 14.85 & 8.80 & 13.01 & 8.87 & 9.90 & 9.47 \\
\cline{2-12}
\cite{krizhevsky2009learning} & avg. & 17.75 & 8.07 & \textbf{15.51} & \textbf{8.34} & 14.62 & 8.68 & 13.13 & 8.88 & \textbf{10.34} & \textbf{9.37} \\
& std. & 0.19 & 0.02 & 0.13 & 0.07 & 0.23 & 0.10 & 0.10 & 0.01 & 0.40 & 0.10 \\
\hline
\end{tabular}

\label{table:cifar}
\end{table*}

\section{Experiments}
\label{sec4}

\subsection{Implementation Details}
\label{sec4.1}
\textbf{Datasets.} To show the superiority of the proposed method, we conduct extensive experiments on the various datasets: CIFAR-10~\cite{krizhevsky2009learning}, CIFAR-100~\cite{krizhevsky2009learning}, LSUN~\cite{yu2015lsun}, and tiny-ImageNet~\cite{yao2015tiny}, which is a subset of ImageNet~\cite{deng2009imagenet}, consisting of the 200 selected classes. Specifically, among the various classes in LSUN dataset, we employ the church and tower images to our experiments. The resolutions of CIFAR-10 and CIFAR-100 datasets are $32\times 32$, whereas images of LSUN and tiny-ImageNet datasets are resized to $128\times 128$ pixels. We use the hinge-version loss as the objective function in all experiments, except for the ablation studies in Table 7 of the main manuscript. In addition, since all parameters in the discriminator and generator including the proposed method can be differentiated, we employ the Adam optimizer~\cite{kingma2014adam} and set the user parameters of Adam optimizer, \ie $\beta _1$ and $\beta _2$, to 0 and 0.9, respectively. 

For training CIFAR-10 and CIFAR-100 datasets, we set the learning rate as 2e-4, and the discriminator was updated 5 times using different mini-batches when the generator is updated once. Also, we set a batch size of the discriminator as 64 and trained the generator for 50k iterations. In our experiments, following the previous papers~\cite{miyato2018cgans, miyato2018spectral, park2021GRB}, the generator is trained with a batch size twice as large as when training the discriminator. That means the generator and discriminator are trained with 128 and 64 batch size, respectively. In contrast, for training LSUN-church, LSUN-tower, and tiny-ImageNet datasets, we employ a two-time scale update rule~(TTUR)~\cite{heusel2017gans} where the learning rates of the generator and discriminator are set to 1e-4 and 4e-4, respectively. In the TTUR technique, the discriminator is updated a single time when the generator is updated once. We set batch sizes of the discriminator and the generator to 32. The network is trained for 300k iterations on the LSUN-church and LSUN-tower datasets, and 1M iterations on the tiny-ImageNet dataset. We reduce the learning rate linearly over the last 50k iterations for all datasets. All models are trained on a single RTX 3090 GPU.
\newline
\newline
\textbf{Network Architecture.} To measure the effectiveness of the proposed method, like the state-of-the-art studies~\cite{miyato2018cgans, brock2018large, zhang2019self, park2021GRB}, we design the generator and discriminator following the strong baselines, \ie SNGAN and BigGAN~\cite{miyato2018spectral, brock2018large}. More specifically, we adopt the generator and discriminator architectures constructing with multiple residual blocks as our baseline models. To train the networks, we replace the conventional normalization techniques of the residual block in the generator with the proposed method. Figure~\ref{sfig1} shows the detailed architectures of the residual blocks in the conventional and proposed methods. The generator and discriminator architectures in StyleGAN~\cite{karras2019style} are also widely used to evaluate GAN performance, but only cover GAN, not cGAN. Therefore, to show the effectiveness of the proposed method in both GAN and cGAN, we decide not to use StyleGAN-based generator and discriminator architectures.


For training CIFAR-10 and CIFAR-100 datasets the spectral normalization (SN)~\cite{miyato2018spectral} is only used for the discriminator, whereas the SN is applied to both generator and discriminator for training LSUN-church, LSUN-tower, and tiny-ImageNet datasets. In the discriminator, the feature maps are down-sampled by utilizing the average-pooling after the second convolution. In the generator, the up-sampling (a nearest-neighbor interpolation) operation is located before the first convolution. Descriptions of the network architectures for the generator and discriminator are described in Table~\ref{table:gnd}. To train the network in the conditional GAN~(cGAN) framework, following the most representative cGAN scheme, we add the conditional projection layer in the discriminator~\cite{miyato2018cgans}.
\newline\newline
\textbf{Evaluation metrics.} We employ the most popular assessments, FID~\cite{heusel2017gans} and IS~\cite{salimans2016improved}, which evaluate how `realistic' the generated image is. In particular, the FID computes the Wasserstein-2 distance between the distributions of the generated and real samples, \ie $P_g$ and $P_r$, in the feature space of the Inception model~\cite{szegedy2016rethinking}. The generated samples with better quality have lower FID. 
On the other hand, the IS measures the KL divergence between the conditional and marginal class distributions. Salimans~\etal~\cite{salimans2016improved} demonstrated that it is strongly correlated with the subjective human judgment of image quality. In contrast with the FID, the better the quality of the generated image, the higher the IS. In our experiments, we randomly generate 50,000 samples and compute the FID and IS using the same number of the real images.

\begin{table}[t]
\caption{Comparison of GAN performance on LSUN dataset in terms of the FID.}
\vspace{2mm}
\centering
\begin{tabular}{c|c||>{\centering}p{1.2cm}|>{\centering\arraybackslash}p{1.2cm}}
\hline
\multirow{2}{*}{Dataset} && SNGAN & \multirow{2}{*}{Proposed} \\
&& \cite{miyato2018spectral} & \\
\hline\hline
& trial1 & 8.16 & 6.72 \\
& trial2 & 7.83 & 7.02 \\
LSUN-church & trial3 & 8.22 & 6.98 \\
\cline{2-4}
\cite{yu2015lsun} & avg. & 8.07 & \textbf{6.91} \\
& std. & 0.21 & 0.16 \\
\hline
& trial1 & 12.42 & 10.52 \\
& trial2 & 12.29 & 11.65 \\
LSUN-tower & trial3 & 12.54 & 11.00 \\
\cline{2-4}
\cite{yu2015lsun} & avg. & 12.42 & \textbf{11.06} \\
& std. & 0.12 & 0.57 \\
\hline
\end{tabular}

\label{table:lsun}
\end{table}

\begin{figure}[t]
\centering
\includegraphics[width=0.9\linewidth]{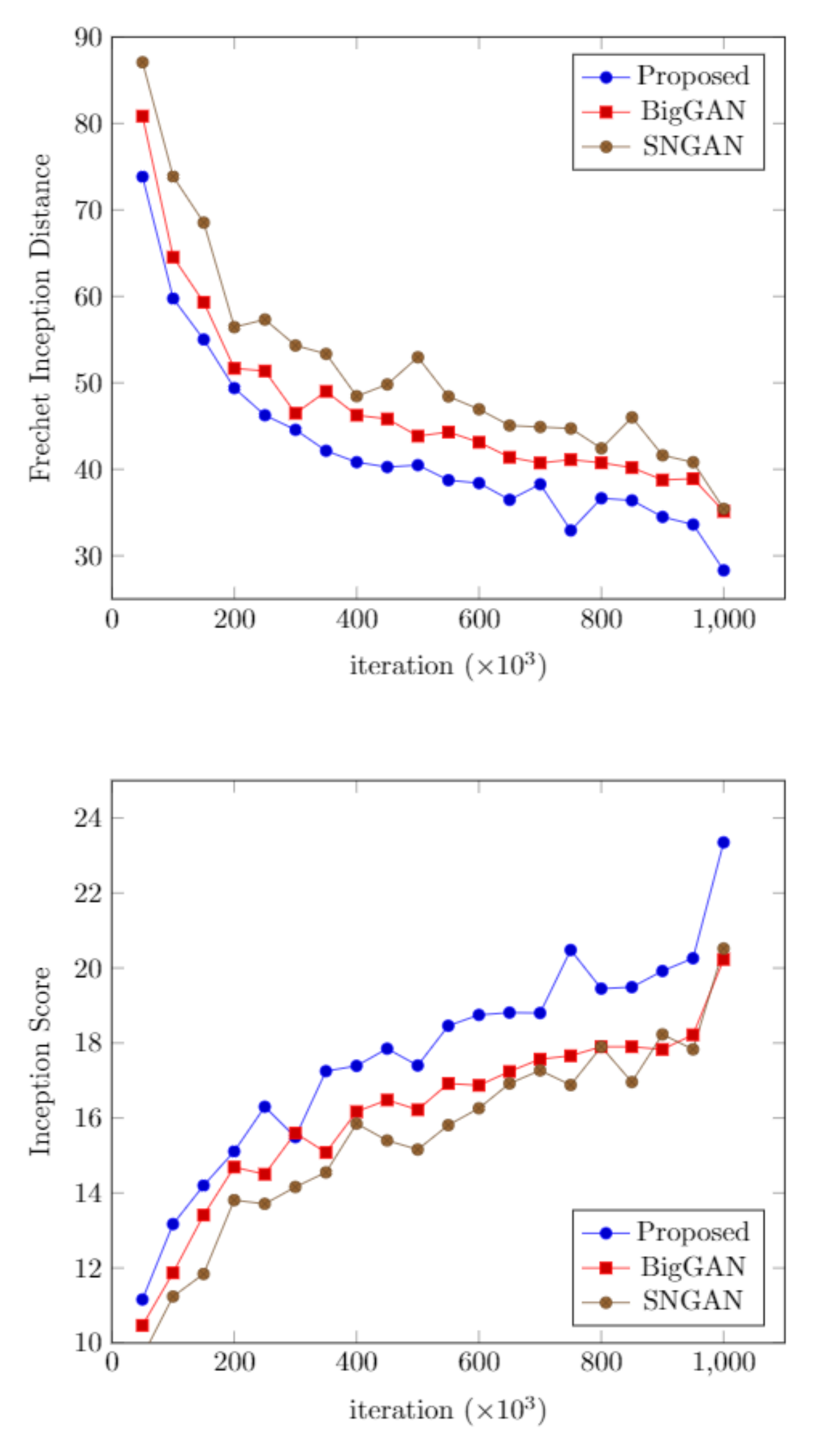}
\caption{The learning curves showing the performance growth of the FID~(top) and IS~(bottom) over the training iteration.}
\label{fig5}
\end{figure}

\begin{table}[t]
\caption{Comparison of cGAN performance on tiny-ImageNet dataset in terms of the FID and IS.}
\vspace{0.3cm}
\centering
\begin{tabular}{>{\centering}p{2.5cm}|>{\centering}p{1.5cm}>{\centering\arraybackslash}p{1.5cm}}
\hline
\multirow{2}{*}{Model} & \multicolumn{2}{c}{tiny-ImageNet~\cite{yao2015tiny, deng2009imagenet}} \\
\cline{2-3}
& FID & IS \\
\hline\hline
SNGAN~\cite{miyato2018spectral} & 35.42 & 20.52 \\
BigGAN~\cite{brock2018large} & 35.13 & 20.23 \\
Proposed & \textbf{28.31} & \textbf{23.35} \\
\hline
\end{tabular}

\label{table:tin}
\end{table}

\begin{figure*}[t] 
\centering
\includegraphics[width=0.9\textwidth]{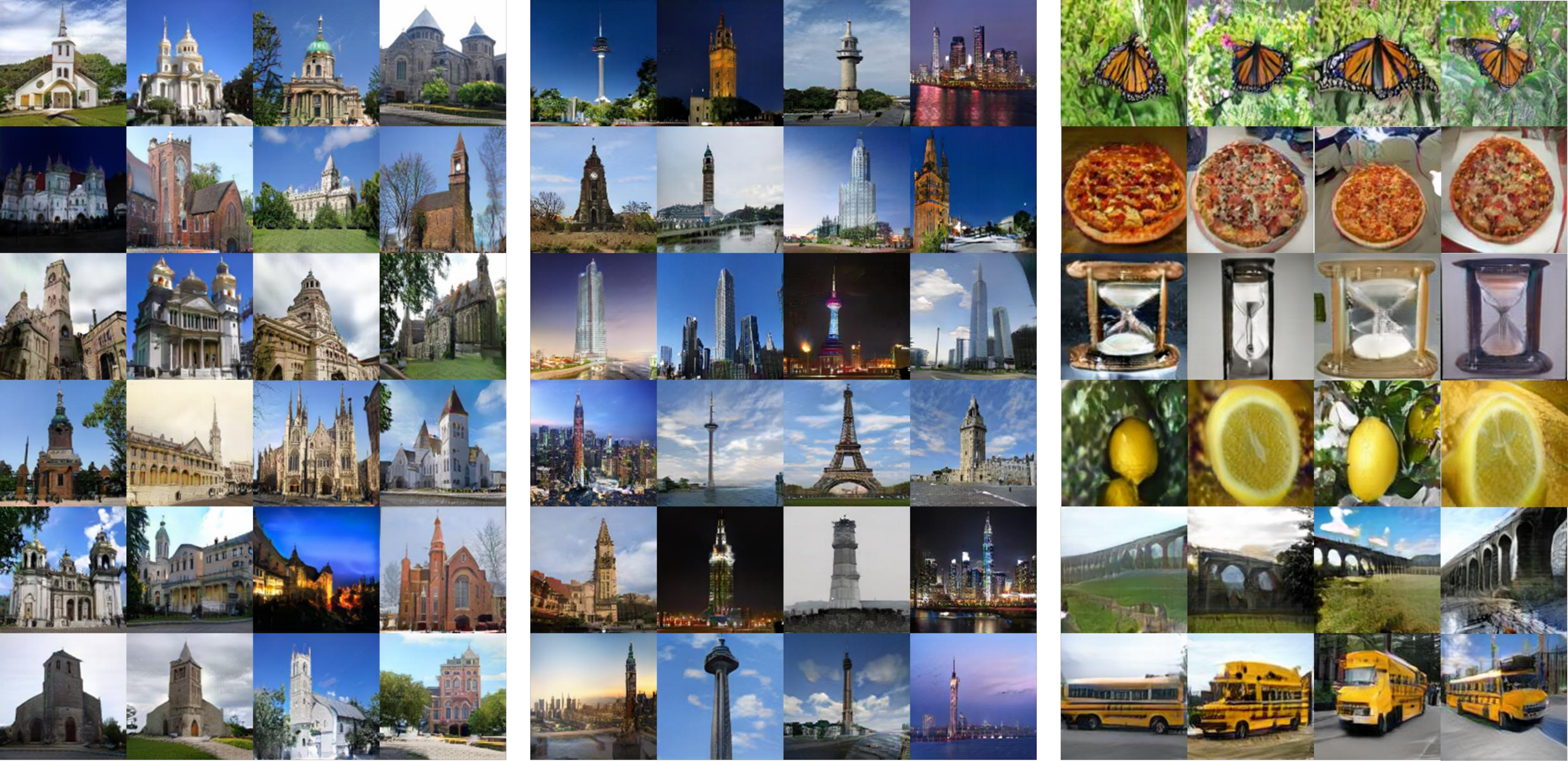}
\caption{Samples of the generated images using the proposed method on LSUN-church, LSUN-tower, and tiny-ImageNet datasets.}
\label{fig6}
\end{figure*}

\subsection{Experimental Results}
\label{sec4.2}
Following the state-of-the-art studies~\cite{miyato2018cgans, brock2018large, zhang2019self}, we employ a strong baseline called spectrally normalized GAN (SNGAN)~\cite{miyato2018spectral}, which is widely used to this day; we only replace the normalization in the generator with the proposed SPN and do not alter the architecture of the discriminator. For cGAN, specifically, BigGAN~\cite{brock2018large} is a SNGAN-like structure in which noise is additionally input to the residual block. Thus, in the cGAN scheme, we compare the performance of the proposed method with those of SNGAN and BigGAN. 

To demonstrate the effectiveness of the proposed SPN, we conduct the preliminary experiments of the image generation task on CIFAR-10 and CIFAR-100 datasets. Table~\ref{table:cifar} summarizes the comprehensive results obtained by applying the conventional and proposed methods, and the bold numbers indicate the best performance among the results. As shown in Table~\ref{table:cifar}, on both CIFAR-10 and CIFAR-100 datasets, SPN consistently achieves the better performance than its counterpart in terms of FID and IS. These results reveal that using SPN is more effective than utilizing BN in the generator. In addition, in the cGAN framework, the proposed method outperforms the current state-of-the-art methods, \ie SNGAN and BigGAN, by a large margin in all the datasets. For instance, on CIFAR-100 dataset, the proposed method achieves the FID of 10.34, which is about 29.27\% and 21.25\% better than SNGAN and BigGAN, respectively. 

\begin{table}[t]
\caption{When adopting BN or SPN, the generative performance on CIFAR-10 dataset and number of network parameters and FLOPs. ${\dagger}$ means a channel-increased model so as to match the computational cost of the proposed method. Here, M/B indicates million/billion ($10^{6}$/$10^{9}$), respectively.}
\vspace{2mm}
\centering
\begin{tabular}{c|c|c|c}
\hline
& SNGAN & SNGAN$^{\dagger}$ & Proposed \\
\hline\hline
FID & 13.46$\pm$0.30 & 13.61$\pm$0.24 & \textbf{12.16$\pm$0.16} \\
IS & 7.77$\pm$0.02 & 7.74$\pm$0.05 & \textbf{7.93$\pm$0.02} \\
\hline
$\#$Param. & 4.07M & 4.53M & 4.52M \\
$\#$FLOPs. & 1.59B & 1.79B & 1.79B \\
\hline
\end{tabular}

\label{table:param}
\end{table}

\begin{table}[t]
\caption{Performance of the proposed method with different configurations on CIFAR-10 dataset. We change the kernel size of the convolution in the affine parameter estimation module and the number of channels of $m$. Also, $*$ indicates the case of employing the standard convolution instead of the depth-wise convolution.}
\vspace{2mm}
\begin{center}
\begin{tabular}{c|c|c|c}
\hline
\multirow{2}{*}{Kernel size} & $\#$Ch. & \multirow{2}{*}{FID} & \multirow{2}{*}{IS} \\
& of $m$ && \\
\hline\hline
$1\times1$ & $c$ & 12.26$\pm$0.29 & 7.82$\pm$0.01 \\
$3\times3$ & $c$ & \textbf{12.16$\pm$0.16} & 7.93$\pm$0.02 \\
$5\times5$ & $c$ & 12.26$\pm$0.35 & \textbf{7.99$\pm$0.03} \\
\hline
$3\times3$ & $1$ & 12.76$\pm$0.28 & 7.88$\pm$0.05 \\
$3\times3$ & $c^{*}$ & 67.09$\pm$8.69 & 5.57$\pm$0.09 \\
\hline
\end{tabular}
\end{center}
\vspace{-0.3cm}

\label{table:ablation}
\end{table}

To ensure the ability of SPN when producing images in challenging datasets, we employ LSUN-church and tower datasets for GAN. As shown in Table~\ref{table:lsun}, the proposed method effectively improves the GAN performance on LSUN-church and tower datasets. Furthermore, to evaluate the effectiveness of conditional SPN, we train the network using the tiny-ImageNet dataset. In our experiments, to show the effectiveness of conditional SPN more reliably, we further present the FID and IS curves exhibiting the performance growth on tiny-ImageNet over the training iteration in Figure~\ref{fig5}. As we can see, the proposed method consistently outperforms the conventional methods, \ie SNGAN and BigGAN, during the training procedure. The final FID and IS values are summarized in Table~\ref{table:tin}. Based on these results, we conclude that the proposed method is effective to significantly improve the performance of cGAN as well as GAN. On the other hand, Figure~\ref{fig6} shows the samples of the generated images. As we can see, the proposed method is effective to synthesize the images with complex scenes. These overall results indicate that SPN and conditional SPN can generate visually pleasing images on challenging datasets.

\subsection{Ablation Studies}
\label{sec4.3}
Indeed, SPN employs slightly more network parameters to build the self-latent mask and to predict the affine transformation parameters. Thus, for SPN to be practically used, it needs to achieve a fine trade-off between improved performance and increased computational cost. To explain the efficiency of SPN clearly, we compare the number of network parameters and floating operations~(FLOPs) of the conventional and proposed methods. As shown in Table~\ref{table:param}, the proposed method replacing 6 BNs of SNGAN with SPNs uses additional parameters of 0.05M and FLOPs of 0.20B; the proposed method marginally increases the computational complexity. One may anticipate that the increased number of computational cost results in improving the performance. To alleviate this issue, we conduct ablation study that equalizes the computational cost of the generator. More specifically, we increase the number of output channels in the generator of SNGAN so as to match the computational requirement. As described in Table~\ref{table:param}, this variation does not contribute to any noticeable performance improvement. Thus, we believe that the performance improvement is caused by the spatially varied transformation, not by the additional computational cost.

\begin{table}[t]
\caption{Performance evaluation of the conventional and proposed methods with/without the bias term on CIFAR-10 (top) and CIFAR-100 (bottom) datasets. The bold and underlined numbers indicate the best and the second-best performance among the results, respectively.}
\vspace{2mm}
\begin{center}
\begin{tabular}{cc|cc}
\hline
Method & Bias & FID & IS \\
\hline\hline
cBN && 10.21$\pm$0.18 & 8.03$\pm$0.03 \\
ccBN & \checkmark & 9.45$\pm$0.15 & 8.03$\pm$0.06 \\
\hline
\multirow{2}{*}{cSPN (ours)} && \underline{7.88$\pm$0.09} & \underline{8.28$\pm$0.02} \\
& \checkmark & \textbf{7.72$\pm$0.18} & \textbf{8.35$\pm$0.06} \\
\hline
\end{tabular}

\vspace{0.5cm}
\begin{tabular}{cc|cc}
\hline
Method & Bias & FID & IS \\
\hline\hline
cBN && 14.62$\pm$0.23 & 8.68$\pm$0.10 \\
ccBN & \checkmark & 13.13$\pm$0.10 & 8.88$\pm$0.01 \\
\hline
\multirow{2}{*}{cSPN (ours)} && \underline{11.06$\pm$0.19} & \underline{9.28$\pm$0.06} \\
& \checkmark & \textbf{10.34$\pm$0.40} & \textbf{9.37$\pm$0.10} \\
\hline
\end{tabular}
\end{center}
\vspace{-0.5cm}

\label{table:noise}
\end{table}

In order to show the generalization ability of SPN, we conduct some ablation studies of the detailed components in the proposed method. Table~\ref{table:ablation} lists the performance of variations of the proposed SPN. First, we vary the convolutional kernel size acting on $m$ and $m^{*}$. It hurts the performance to use a kernel size of $1\times1$ since it cannot utilize the local context information of the masks. Meanwhile, a kernel size of $5\times5$ shows similar performance to the $3\times3$ kernel. Thus, in the rest of our experiments, we adopt the kernel size of $3\times3$ for computational efficiency. On the other hand, to show the validity of producing an independent mask for each channel, we measure the performance of the model trained with \textit{m} having a single channel. As summarized in Table~\ref{table:ablation}, the single channel mask 
degrades performance since it has insufficient variations to derive the proper transformation parameters for each channel, \ie $\gamma(j)$ and $\beta(j)$. In addition, instead of using the depth-wise convolution, we apply the standard convolution to $m$ having $c$ channels to estimate the transforming parameters. However, this model fails to train since there are too many parameters in the normalization layer; it is hard to train stably because of an imbalance between the generator and discriminator.

\begin{table}[t]
\caption{Performance evaluation in different loss settings on CIFAR-10 dataset.}
\vspace{0.1cm}
\begin{center}
\begin{tabular}{c|c|c|c}
\hline
Loss && \multirow{2}{*}{SNGAN} & \multirow{2}{*}{Proposed} \\
function &&& \\
\hline\hline
\multirow{2}{*}{CE~\cite{goodfellow2014generative}} & FID & 17.41$\pm$0.64 & \textbf{15.13$\pm$1.25} \\
& IS & 7.52$\pm$0.02 & \textbf{7.65$\pm$0.11} \\
\hline
\multirow{2}{*}{LSGAN~\cite{mao2017least}} & FID & 20.57$\pm$0.83 & \textbf{18.74$\pm$0.45} \\
& IS & 7.17$\pm$0.02 & \textbf{7.42$\pm$0.04} \\
\hline
\end{tabular}
\end{center}
\vspace{-0.3cm}

\label{table:loss}
\end{table}

\begin{table}[t]
\caption{Performance comparison of the conventional and proposed methods with/without the spatial attention (SA) technique on CIFAR-10 dataset.}
\vspace{0.1cm}
\begin{center}
\begin{tabular}{cc|cc}
\hline
\multirow{2}{*}{Model} & SA & \multirow{2}{*}{FID} & \multirow{2}{*}{IS} \\
& \cite{woo2018cbam} && \\
\hline\hline
\multirow{2}{*}{SNGAN} && 13.46$\pm$0.30 & 7.77$\pm$0.02 \\
& \checkmark & 13.67$\pm$0.24 & 7.76$\pm$0.04 \\
\hline
\multirow{2}{*}{Proposed} && 12.16$\pm$0.16 & 7.93$\pm$0.02 \\
& \checkmark & \textbf{11.73$\pm$0.17} & \textbf{7.99$\pm$0.03} \\
\hline
\end{tabular}
\end{center}
\vspace{-0.3cm}
\label{table:sa}
\end{table}

In the conditional SPN~(cSPN), we employ the noise vector to produce the bias term for $\gamma_c(j)$ and $\beta_c(j)$. To show the effectiveness of the bias term, we conduct ablation studies that compare the performance of the network trained with/without the bias term. The experimental results are summarized in Table~\ref{table:noise}. As we can see, the bias term slight improves the performance of the proposed cSPN on both CIFAR-10 and CIFAR-100 datasets. These results indicate that the bias term derived from the latent space can encourage the performance improvement. It is worth noting that the proposed method without the bias term still outperforms the conventional methods, \ie cBN and ccBN. That means even someone builds the generator that is difficult to provide the bias term to cSPN, the network would achieve the better performance than the existing normalization methods.

Furthermore, we conduct additional experiments in which the networks are trained with two different objective functions: the function based on the cross-entropy (CE) theorem and the function proposed in the least square GAN (LSGAN) paper~\cite{mao2017least}. As shown in Table~\ref{table:loss}, the proposed method exhibits the performance improvement steadily even with various loss functions. These results reveal that the proposed method can be effectively applied to the conventional GAN without considering experimental settings such as the adversarial loss function.

One might perceive that the proposed method is similar to the technique that directly feeds spatial attention to the feature map. However, while the existing spatial attention methods are additionally applied after the convolutional layer, SPN is an improved normalization algorithm. Therefore, the proposed method is a differentiated technique that can be used together with the spatial attention method. To prove this claim, we conduct experiments that train the network with/without the well-known spatial attention~(SA) technique~\cite{woo2018cbam}. As presented in Table~\ref{table:sa}, the generator trained with SA shows analogous performance to that trained without SA. In contrast, the proposed method slightly improves the generative performance when using SA. Based on these results, we expect that further improvement can be achieved by designing a new SA technique fitting well with the proposed method.
\vspace{-0.25cm}

\begin{figure}[t] 
\centering
\includegraphics[width=1.0\linewidth]{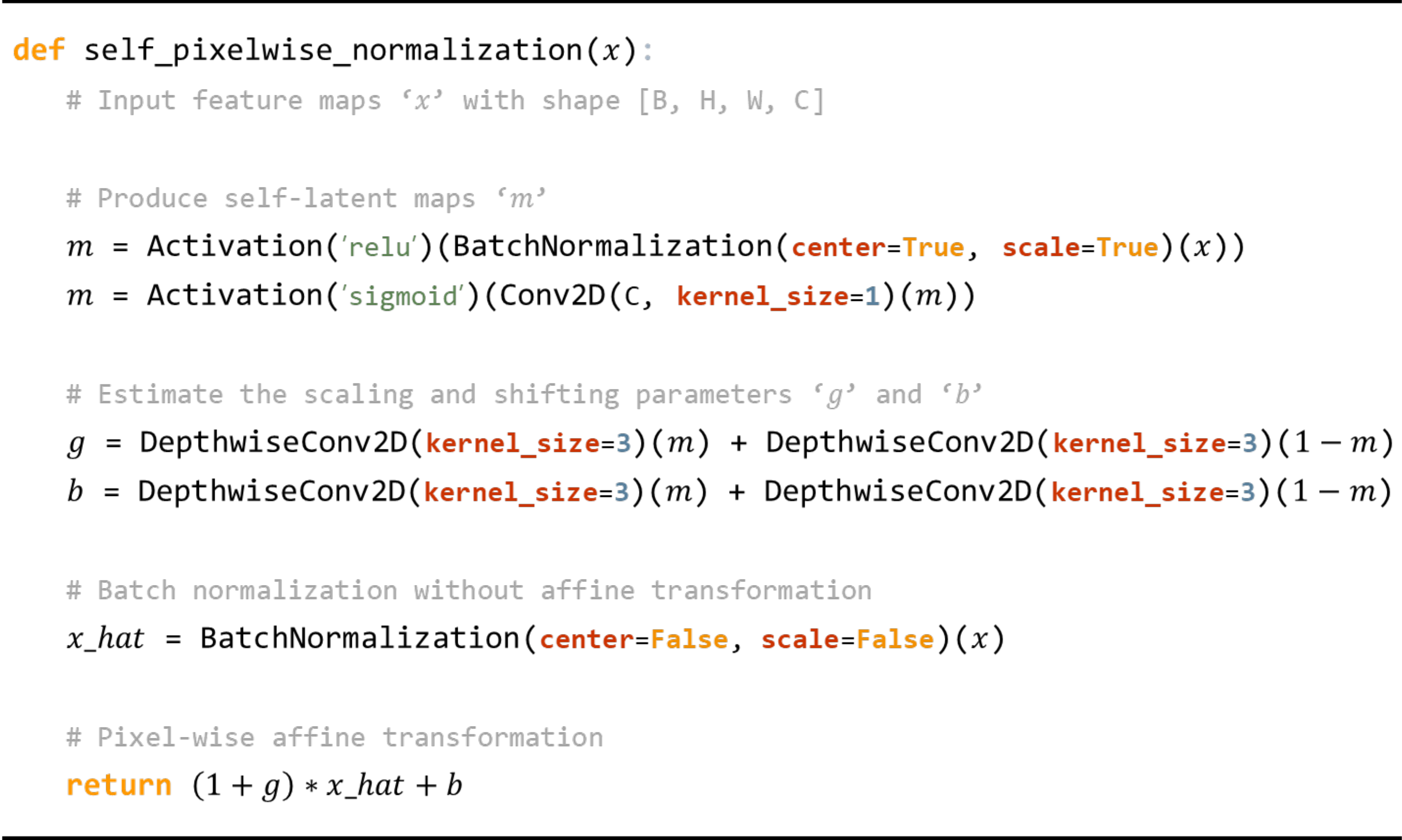}
\caption{Python code of the proposed SPN. This is implemented in the \textit{Tensorflow} and \textit{Keras} API.}
\label{sfig2}
\vspace{-0.25cm}
\end{figure}

\section{Analysis and Discussion}
\label{sec5}

\textbf{Why does SPN work better?} A short answer to this question is that SPN is better to provide the region-specific features than the conventional normalization methods. Indeed, while BN~\cite{ioffe2015batch} is an essential piece in almost all generators, it cannot embed the spatially varied information since it employs the same scaling and shifting parameters for all pixels. In other words, when using BN, the convolutional layer should learn the region-specific features itself. However, this work is quite taxing on the convolutional layer due to the absence of the external guidance. In contrast, since SPN allows the intrinsic pixel-wise affine transformation, the proposed model could produce the spatially adaptive features in the normalization procedure. That means it can offload the burden of the convolutional layer. Therefore, SPN and convolutional layer have a synergistic effect to derive the features appropriate for each region, which can remarkably improve the generative performance. We present the simply abbreviated code for SPN on the \textit{Tensorflow}~\cite{tensorflow2015-whitepaper} platform. The code using the \textit{Keras} API~\cite{chollet2015keras} is presented in Figure~\ref{sfig2}.

Furthermore, we theoretically analyze the proposed method. In fact, SPN is related to and generalized to several existing normalization methods. First, when replacing $m(j)$ with the spatially-invariant constant value, we arrive at the form of BN since all values in $\gamma(j)$ or $\beta(j)$ are same. Similarly, we can make the form of LN~\cite{ba2016layer} by replacing $m$ with the scalar value. Unlike the previous normalization methods in which some or all of $m$ is uniform, the proposed SPN generates the spatially adaptive $m$ for each input feature map. Again, $m$ is an internally produced \textit{self-latent mask} for the feature itself. Hence, the proposed method is more suitable for the image generation.
\newline \newline
\textbf{Limitations.} Despite the significant improvements, SPN still faces with one confusion: the propsoed method is designed with an assumption that the generated images contain a single object and background. That means we do not consider applying SPN to the image-to-image task that synthesizes the images having multiple objects with different classes. Thus, we believe that future investigation should be required to employ SPN to the image-to-image translation technique.
\vspace{-0.5cm}

\section{Conclusion}
\label{sec6}

In this paper, we introduce a novel normalization technique, called SPN, adopted for the generator of both GAN and cGAN schemes. The proposed method performs the pixel-adaptive affine transformation with only feature-intrinsic data. Without a complex modification, our technique is simply applied to the existing models by replacing the conventional normalization. We validate the superiority of SPN with visualization of its self-latent masks and the extensive experimental results. In addition, we further investigate the proposed method in its broad aspects with ablation studies. Therefore, it is expected that our method is applicable to various GAN-based applications.

\bibliographystyle{spmpsci}      
\bibliography{egbib.bib}

\vspace{30mm}

\begin{wrapfigure}{l}{0\textwidth}
\includegraphics[width=1in, height=1.25in]{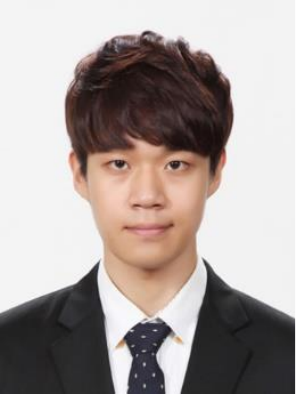}
\end{wrapfigure} 
\vspace{5mm}
\textbf{Yoon-Jae Yeo} received his B.S. degree in Electrical Engineering from Korea University in 2017. He is currently pursuing his Ph.D. degree in Electrical Engineering at Korea University. His research interests are in the areas of image processing, computer vision, and deep learning.\newline

\begin{wrapfigure}{l}{0\textwidth}
\includegraphics[width=1in, height=1.25in]{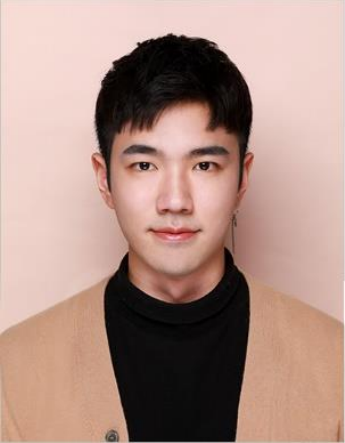}
\end{wrapfigure} 
\vspace{5mm}
\textbf{Min-Cheol Sagong} received his B.S. degree in Electrical Engineering from Korea University in 2018. He is currently pursuing his Ph.D. degree in Electrical Engineering at Korea University. His research interests are in the areas of digital signal processing, computer vision, and artificial intelligence.\newline

\begin{wrapfigure}{l}{0\textwidth}
\includegraphics[width=1in, height=1.25in]{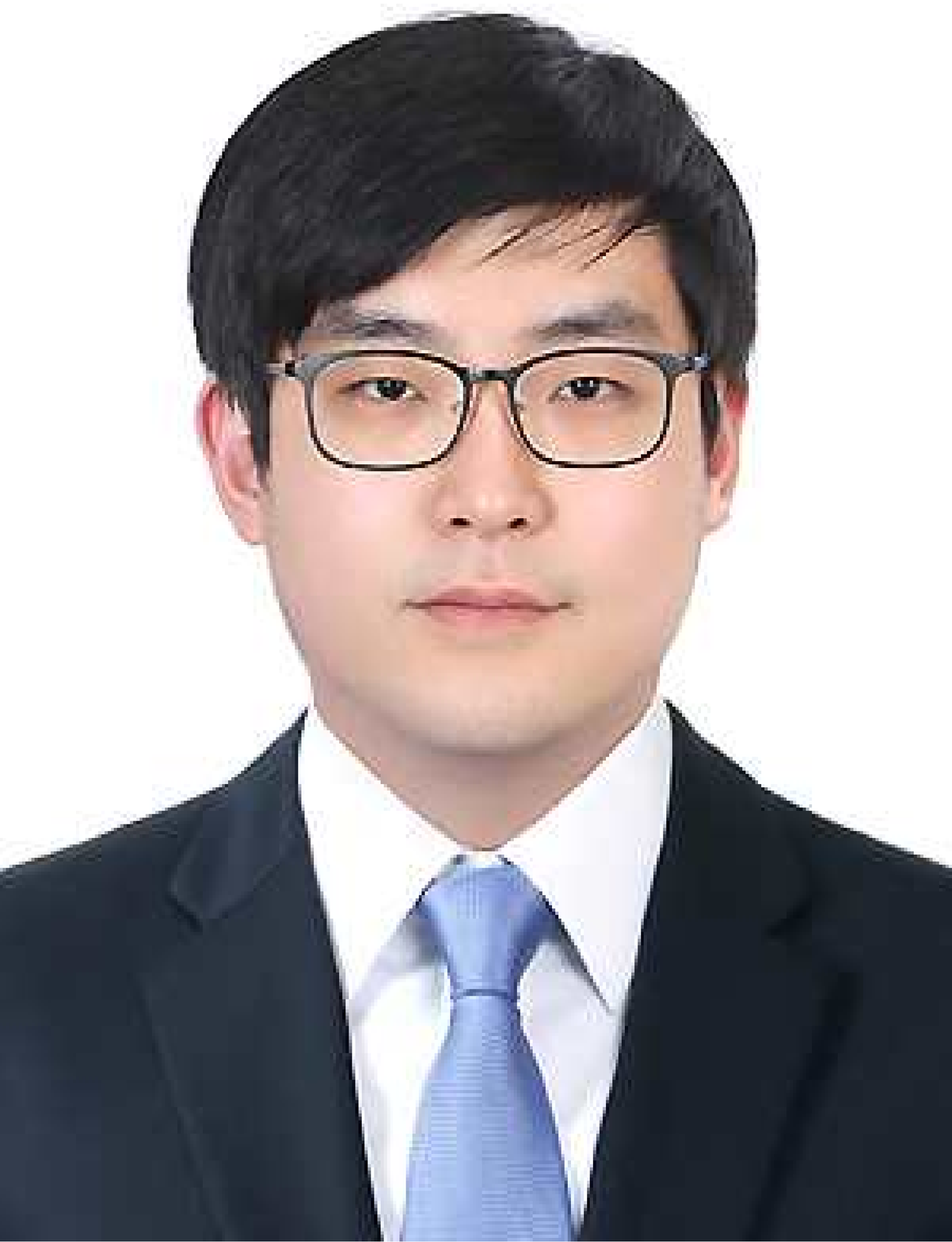}
\end{wrapfigure} 
\vspace{5mm}
\textbf{Seung Park} received the B.S. and Ph.D. degrees in electrical engineering from Korea University, Seoul, South Korea, in 2013 and 2020, respectively. He is currently a Clinical Assistant Professor in Chungbuk National University Hospital. His current research interests include computer vision and image processing. \newline

\begin{wrapfigure}{l}{0\textwidth}
\includegraphics[width=1in, height=1.25in]{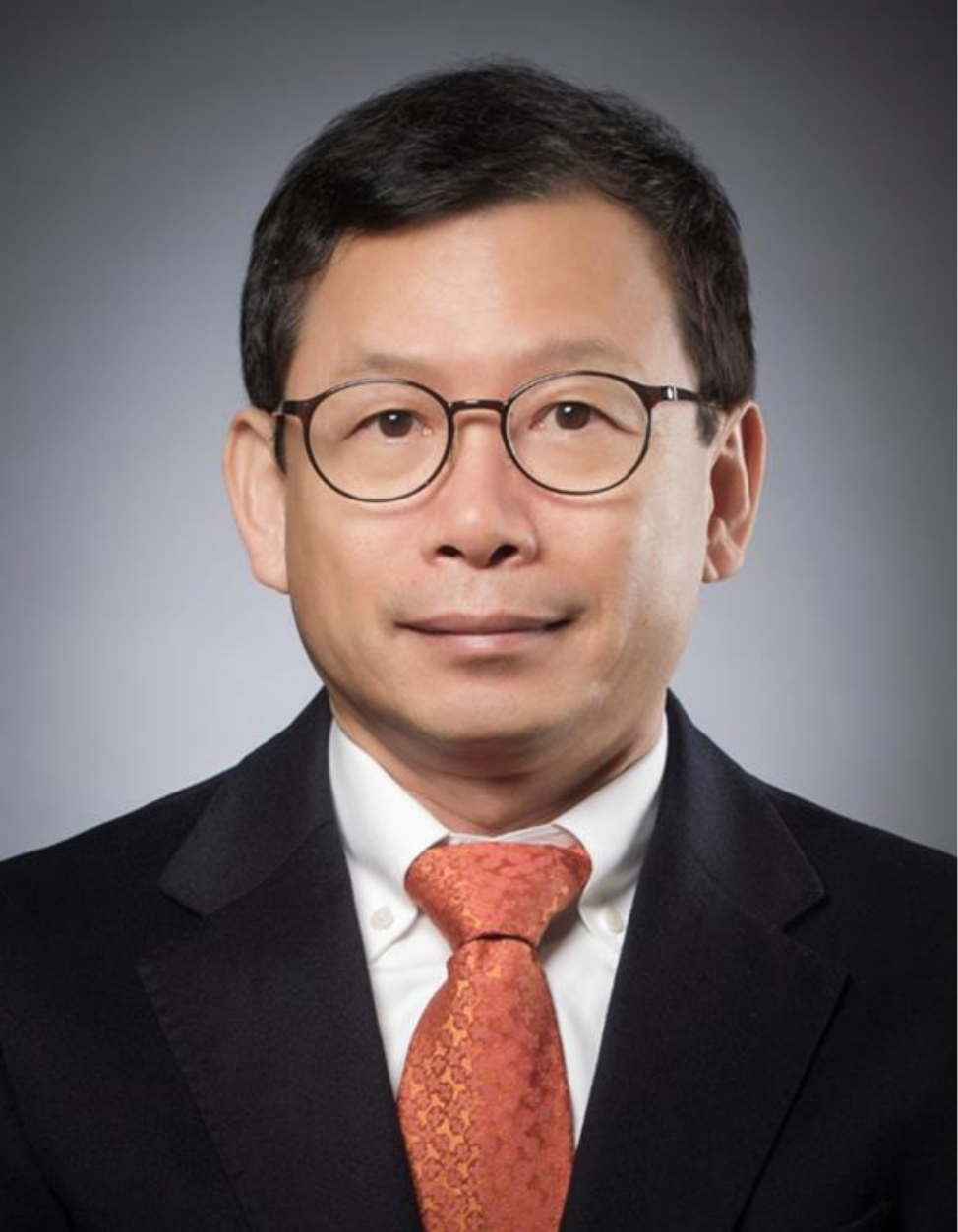}
\end{wrapfigure} 
\vspace{5mm}
\textbf{Sung-Jea Ko} (M’88-SM’97-F’12) received his Ph.D. degree in 1988 and his M.S. degree in 1986, both in Electrical and Computer Engineering, from State University of New York at Buffalo, and his B.S. degree in Electronic Engineering at Korea University in 1980. In 1992, he joined the Department of Electronic Engineering at Korea University where he is currently a Professor. From 1988 to 1992, he was an Assistant Professor in the Department of Electrical and Computer Engineering at the University of Michigan-Dearborn. He has published over 210 international journal articles. He also holds over 60 registered patents in fields such as video signal processing and computer vision. 

Prof. Ko received the best paper award from the IEEE Asia Pacific Conference on Circuits and Systems (1996), the LG Research Award (1999), and both the technical achievement award (2012) and the Chester Sall award from the IEEE Consumer Electronics Society (2017). He was the President of the IEIE in 2013 and the Vice-President of the IEEE CE Society from 2013 to 2016. He is a member of the National Academy of Engineering of Korea. He is a member of the editorial board of the IEEE Transactions on Consumer Electronics.

\begin{wrapfigure}{l}{0\textwidth}
\includegraphics[width=1in, height=1.25in]{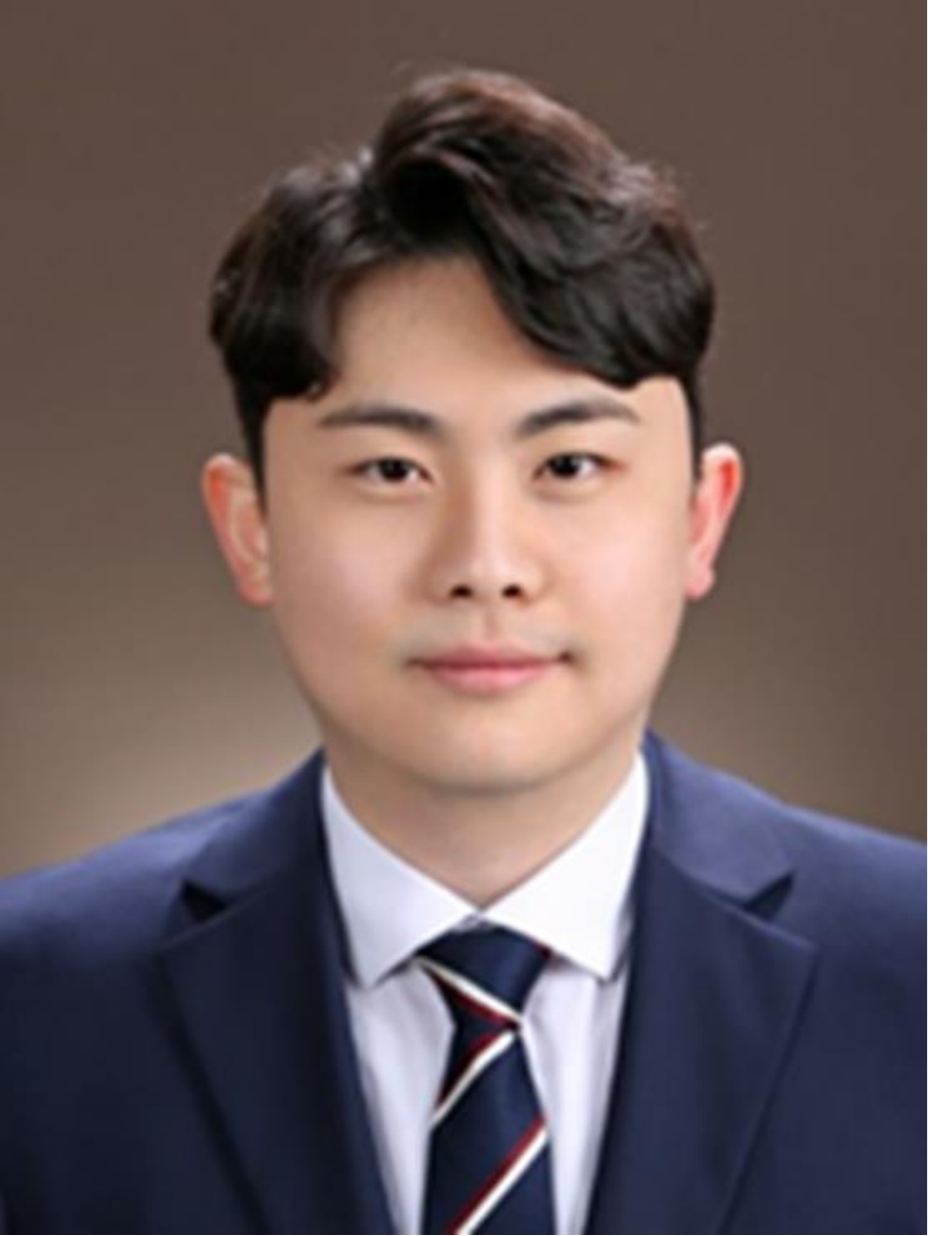}
\end{wrapfigure} 
\vspace{5mm}
\textbf{Yong-Goo Shin} received the B.S. and Ph.D. degrees in electronics engineering from Korea University, Seoul, South Korea, in 2014 and 2020, respectively. He is currently an Assistant Professor with the Division of Smart Interdisciplinary Engineering, Hannam University. He has published over 10 international journal articles in fields such as image processing, computer vision, and deep learning.



\end{document}